\pdfoutput=1

\documentclass{article}
\usepackage[verbose=true,letterpaper]{geometry}
\newgeometry{
textheight=9in,
textwidth=5.5in,
top=1in,
headheight=12pt,
headsep=25pt,
footskip=30pt
}

\usepackage{xcolor}
\usepackage{times}
\usepackage{latexsym}
\usepackage[T1]{fontenc}
\usepackage[utf8]{inputenc}
\usepackage{microtype}
\usepackage{inconsolata}
\usepackage{graphicx}
\usepackage{wrapfig}
\usepackage{amsmath}
\usepackage{float}
\newtheorem{theorem}{Theorem}
\newtheorem{lemma}{Lemma}
\usepackage{adjustbox}
\usepackage{amsfonts}
\usepackage{url}
\usepackage{hyperref}
\usepackage[numbers]{natbib}
\author{%
$^{a}$Kangjun Noh\\
\and
$^{a}$Baekryun Seong\\
\and
$^{b}$Hoyoon Byun\\
\and
$^{b}$Youngjun Choi\\
\and
$^{c}$Sungjin Song\\
\and
$^{b}$Kyungwoo Song\\
}
\date{%
$^{a}$University of Seoul, Seoul, South Korea
\quad
$^{b}$Yonsei University, Seoul, South Korea
\quad
$^{c}$Crepass Solution, Seoul, South Korea
}

\title{LBC: Language-Based-Classifier for Out-Of-Variable Generalization}

\begin{document}
\maketitle

\begin{abstract}
\noindent Large Language Models (LLMs) have great success in natural language processing tasks such as response generation. However, their use in tabular data has been limited due to their inferior performance compared to traditional machine learning models (TMLs) such as XGBoost. We find that the pre-trained knowledge of LLMs enables them to interpret new variables that appear in a test without additional training, a capability central to the concept of Out-of-Variable (OOV). From the findings, we propose a Language-Based-Classifier (LBC), a classifier that maximizes the benefits of LLMs to outperform TMLs on OOV tasks. LBC employs three key methodological strategies: 1) Categorical changes to adjust data to better fit the model's understanding, 2) Advanced order and indicator to enhance data representation to the model, and 3) Using verbalizer to map logit scores to classes during inference to generate model predictions. These strategies, combined with the pre-trained knowledge of LBC, emphasize the model's ability to effectively handle OOV tasks. We empirically and theoretically validate the superiority of LBC. LBC is the first study to apply an LLM-based model to OOV tasks. The source code is at \textcolor{blue}{\url{https://github.com/sksmssh/LBCforOOVGen.git}}.
\end{abstract}

\begin{figure}[h]
    \centering
    \includegraphics[width=0.6\textwidth]{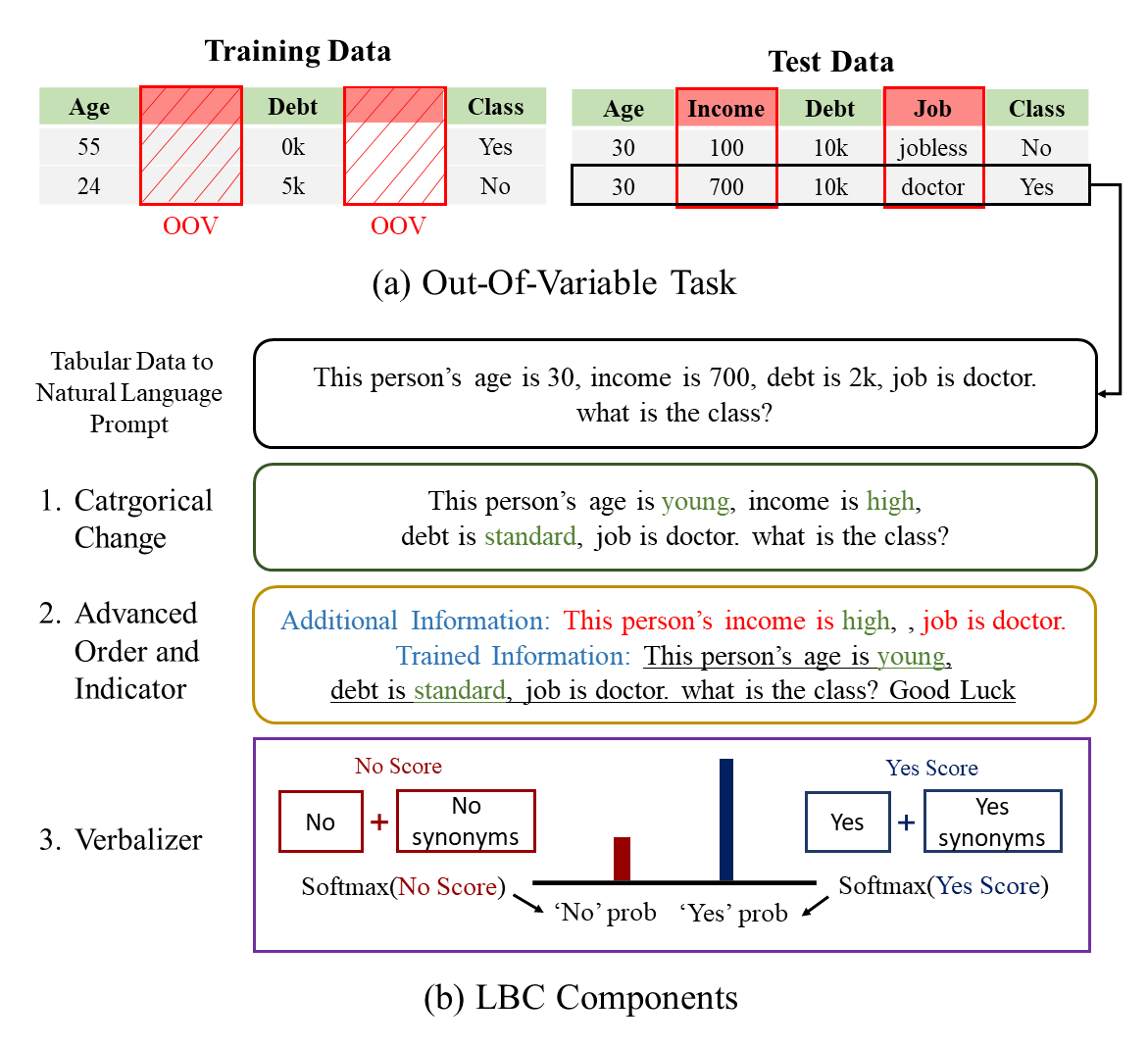}
    \caption{(a) Illustration of OOV task. The variables that were not present in the training data appear in the test data. (b) Key components of LBC to increase performance in OOV tasks. Categorical change refines data to make it easier for LBC to interpret. The advanced order and indicator method enhances the prompts that feed into LBC. The verbalizer aggregates the probabilities for a particular class scattered throughout the logit score and maps them to a specific class.}
    \label{intro}
\end{figure}

\section{Introduction}
LLMs~\cite{gpt1,gpt3,gpt-j,bert} have recently been applied to tabular data. Language-Interfaced-Fine-Tuning (LIFT)~\cite{lift} demonstrated that LLMs achieve reasonable performance on tabular data tasks while maintaining LLM's original structure. However, the pre-trained knowledge of LLMs holds even another potential, their ability to interpret OOV. So, we propose a new model called a \textbf{Language-Based-Classifier (LBC)} to solve the OOV tasks.

OOV tasks are an important problem and are the subject of several ongoing studies~\cite{guo2024outofvariable, DBLP:journals/corr/TzengHDS15, Dreher_2023}. However, studies applying LLM to tabular data do not handle tabular data in an OOV setting. In real-world settings, a variety of constraints often hinder model training, emphasizing the importance of OOV tasks. For example, in healthcare, privacy and regulatory barriers prevent data sharing between hospitals. A model trained on Hospital A's data may encounter new, unseen variables when applied to Hospital B's data, leading to OOV situations. We argue that LBC has strengths in handling OOV tasks, and our rationale is as follows. Converting tabular data to natural language prompts is intuitive, flexible, and easy. This transformation significantly simplifies the handling of OOVs, allowing us to seamlessly handle variables that might not have been discovered during training, overcoming a common limitation of TMLs. Furthermore, LBC leverages the pre-trained knowledge built into LLMs. Unlike TMLs, which struggle with data points or scenarios not present in the training set, LLMs leverage their inherent knowledge. We verified that LBC use OOVs to increase the probability of the correct answer class based on pre-trained knowledge. These advantages are highlighted by the following three methodologies of LBC. First, Categorical Change involves converting numerical types to categorical types like 'high' and 'low' because these variables align better with the LBC, especially in OOV scenarios. Second, Advanced Order and Indicator optimizes the sequence of variables to generate more effective prompts, and introduces indicators to further boost performance. Third, the Verbalizer focuses on mapping LLM's logit scores to the desired class scores, rather than relying on inconsistent output text, improving classification performance. We use the LOw-Rank Adaptation (LoRA)~\cite{lora} to fine-tune the classifier. We theoretically prove that our model approximates an arbitrary classifier with LoRA fine-tuning.

To the best of our knowledge, LBC is the first study to apply an LLM-based classifier to solve the OOV tasks, and we validate LBC's superiority empirically and theoretically.

\section{Related Works}

\subsection{Tabular Data Analysis with LLMs}
LLMs now extend to analyzing tabular data. LIFT~\cite{lift} converts tabular data into natural language prompts for use in LLM, and performs similarly to traditional models like XGBoost~\cite{xgboost}. Models like TP-BERTa~\cite{yan2024makingpretrainedlanguagemodels} and TabPFN~\cite{tabpfn} follow the LM structure but lack the ability to contextualize OOVs. On the other hand, LBC excels at handling OOV tasks and consistently outperforms existing models. LBC's performance in tabular data classification has been validated through theoretical analysis and statistical tests.

\subsection{Out-of-Variable}
Machine learning (ML) models often face the challenge of adapting to new environments with additional, unobserved variables. MomentLearn~\cite{oov-gen} was proposed to address this by using a predictor trained in a source environment and a additional objective matrix for partial derivatives for OOV tasks. However, its application in real-world scenarios is limited. The LBC method overcomes these limitations by leveraging the extensive prior knowledge in LLMs and methodologies for OOV tasks.
Unlike MomentLearn, which is restricted to simple models such as linear or polynomial structures, LBC’s use of LLMs allows for application to more complex models. This enhances its ability to discover intricate relationships between variables and offers greater generalization.
Moreover, MomentLearn’s reliance on an additional matrix, which must be trained with In-Variables, becomes less stable as the ratio of OOVs increases. In contrast, LBC only requires the training of a single predictor and has demonstrated robustness across varying OOV ratios, making it a more efficient and reliable solution for OOV challenges.

\subsection{Verbalizer}

Verbalizer is a mechanism for mapping the various output forms from an LLM to specific classes~\cite{verbalizer,verbalizer2}. Verbalizer contributed to reducing subjective bias in LLM by using a knowledge base to leverage diverse and comprehensive label words. It is also said that the noise of label words in classification can also be improved with a verbalizer. We argue that even in tabular data classification, we need a particular way to map the output of an LLM to the output of a classifier and that we should apply a verbalizer to it.

\subsection{Low-Rank Adaption}
LoRA~\cite{lora} has emerged as an innovation in adapting pre-trained models to specific tasks without extensive retraining of the entire model. LoRA introduces an approach to fine-tuning large pre-trained models. Instead of updating the whole parameter set, LoRA modifies a small subset of the model's weights through a low-rank matrix. This method allows pre-trained models to adapt efficiently while maintaining their original structure and strengths.We theoretically validate the strong classification performance of LBC fine-tuned with LoRA, backed by the proven generalization ability of LoRA~\cite{expressive-power-of-lora}.

\section{Preliminary}

\subsection{Basic Dataset Conversion}
This section describes the process of converting tabular data into prompts for input to LBC. Since our model relies on a frozen pre-trained LLM, converting tabular data into prompts is a crucial step. Let an instance of tabular data with \( K \) features be represented as \([\left[V_1 : x_1\right] , \left[V_2 : x_2\right], \ldots, \left[V_K : x_K\right], \left[\text{class} : y\right]]\), where \(V_k\) is the \(k\)th variable name and \(x_k\) is the \(k\)th variable value.
We need a method for the LLM to clearly distinguish between the variables in this dataset as prompts and the class as the output. This involves creating a conversion technique that clearly marks the end of the prompt and the beginning of the response while ensuring that the answer isn't overly lengthy. Therefore, we format the conversion as: 
``prompt: \(V_1\) is \(x_1\), \(V_2\) is \(x_2\), \ldots, \(V_K\) is \(x_K\). What is the class? label: \(y\)\texttt{@@@}``.

In this setup, the 'prompt' is the input to LLM, and the 'label' is the label for the data instance.
\label{tab:basic_conversion}

\subsection{The Order of the Variables}
During the tabular data to the prompt conversion process, different prompts are generated depending on the variable order.
One instance of tabular data converts to several different types of prompts based on the order of the variables. The total number of prompts that can be generated by changing the order of the variables is $K!$. Every prompt is a transformation of a single instance of tabular data, but the order of the variables gives it a different form, which causes LBC to interpret it differently. Therefore, the order of the variables is a factor that directly affects LBC's performance.
\label{tab:def_order}

\begin{figure*}[]
\centering
\includegraphics[width = \textwidth]{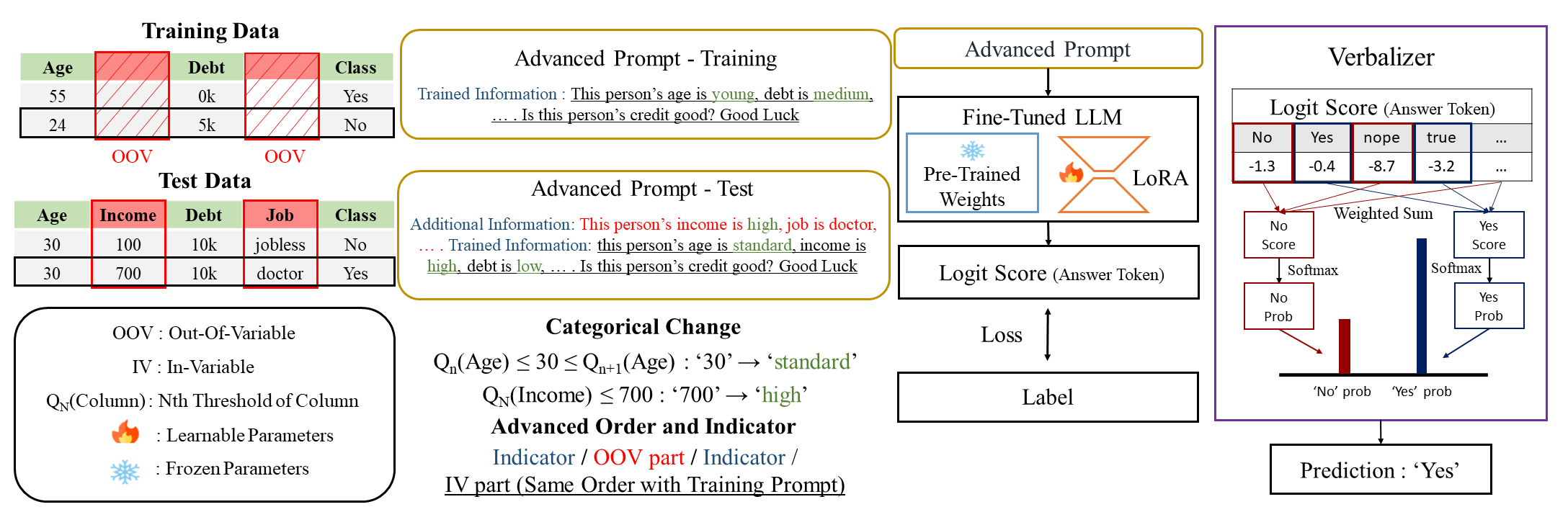}
\caption{The overall process of an LBC performing an OOV task. LBC transforms tabular data into advanced prompt (AP) utilizing strategies that are 1) Categorical change and 2) Advanced order and indicator. These APs are then input into an LLM that has been fine-tuned with a LoRA adapter, to derive a logit score for the answer token. This logit score is assessed against the label to calculate loss, and during inference, the model prediction is generated by mapping the logit score to a class via a 3) Verbalizer.}
\label{fig:main_method}
\vspace{-1.5em}
\end{figure*}

\subsection{Fine-tuning LLM}
Feeding the converted prompts into LBC yields a vector of vocabulary sizes, which is logit for each word in the vocabulary. We use this logit to fine-tune the LLM. Let $\mathbf{Logit}$ be the logit vector for a single input vector.
During fine-tuning, $J$ obtained from the model is used to compute the loss against the true labels. Let $\mathbf{Label}$ be the one-hot encoded vector of the true label for the input. The loss is calculated using a loss function $J$ defined as: \[
    J(\mathbf{Logit}, \mathbf{Label}) = \text{CE}(\mathbf{Logit}, \mathbf{Label})
\]

\noindent where CE is cross-entropy with logit loss function. After calculating the loss, the model's parameters are updated using an optimizer through gradient descent. The update rule in gradient descent can be described as follows: \[
\theta \leftarrow \theta - \eta \nabla_{\theta}J
\]
where $\theta$ is the model's parameters, $\eta$ is the learning rate, and $\nabla_{\theta}J$ is the gradient of the loss with respect to the model parameters.

\subsection{LLM-based Tabular Prediction}
TMLs face significant challenges when processing textual data within feature sets. Text preprocessing inevitably leads to semantic information loss. Despite applying specialized techniques such as one-hot encoding or text vectorization methods (e.g., TF-IDF, Word2Vec, etc.), TMLs remain vulnerable to noise due to its lack of linguistic comprehension. Furthermore, the high dimensionality of text embeddings often impedes efficient learning, and attempts to mitigate this through dimensionality reduction techniques risk further information loss.

In contrast, LLMs offer a promising alternative for handling textual and numerical data in machine learning tasks. LLMs demonstrate superior capability in comprehending semantic content and discerning inter-feature relationships, which is beneficial when critical information is presented textually. 

The previous approaches to LLM-based tabular data classification tasks~\cite{lift} rely on directly comparing the output text generated by the model with class texts such as 'no' or 'yes'. If the prediction is an exact match, it is classified with the corresponding class text. Conversely, if the output text differs, the model's prediction is marked as 'None' and automatically classified as incorrect. To address this limitation, we utilize the logit score to map directly to a specific class rather than using the model's output texts. For this mapping process, we utilize the probability values of the synonyms of the logit score's class text.

\section{Methodology}

\subsection{Categorical Change}
We find that LBC has a better interpretation of categorical variables than numerical ones because it is an LLM-based model. However, this poses a challenge as many key variables in tabular data are numerical. In particular, when LBC deals with OOVs, if the value of the input is numeric, pre-trained knowledge cannot be utilized, unlike categorical type values where the word itself has meaning. Therefore, we need a method to convert numerical variables to categorical types so that LBC leverages its pre-trained knowledge of important variables or OOVs for easier interpretation, and we find that mapping numerical variables to categorical variables using \(N\) categories improves performance. The \(N\) categories are determined based on the principles of \(N\)-tiles, similar to quartiles but dividing the dataset into \(N\) equal parts. The thresholds are the values that divide the dataset into these \(N\) parts. For example, we converted values below the first threshold (\(Q_1\)) to "Category 1", between \(Q_1\) and \(Q_2\) to "Category 2", and so on, up to values above the last threshold (\(Q_{N-1}\)), which are converted to "Category \(N\)". A specific example sentence of Categorical Change is shown in figure~\ref{fig:main_method}.

\subsection{The Advanced Order and Indicator}
As shown in \ref{tab:def_order}, for a single instance of data, different prompts are generated depending on the order of the variables. The same problem occurs in the OOV task, where the number of variables increases due to the addition of OOVs, resulting in more variability in the prompts. This hinders LBC's ability to learn the relationships between tokens. Therefore, we find the format that performs best with optimal learning and inference among a large number of prompt formats, which can vary depending on the order of the OOVs and the trained variables that are not OOVs, called In-Variables (IVs). The format of the training and test prompts with both methods is as follows.

\small{
\begin{flalign*}
&\text{Training Prompt: }\text{IV Indicator} + \underline{\text{IV part}} + \text{Question} \\
&\text{Test Prompt: } \text{OOV Indicator} + \text{OOV part} + \text{IV Indicator} \\
& + \ \text{\underline{IV part}} + \text{Question}
\end{flalign*}
}

\normalsize
\noindent By positioning the OOV part at the front of the prompt and matching the variable order of the \underline{IV part} exactly as in training, the \underline{IV part} in the test prompt has the exact same structure as the \underline{IV part} in the training prompt. This allows LBC to apply the relationships between variables captured during training to the test as well. 
Also, since the indicator is always fixed in the same position, it allows LBC to distinguish between the OOV part and the IV part in training and inference.
A prompt with both categorical change and advanced order and indicator applied is referred to hereafter as an \textbf{advanced prompt (AP)}. An example of an AP can be found in figure~\ref{fig:main_method}.

\subsection{Generalization Ability of LBC: LoRA}
According to ~\citet{expressive-power-of-lora}, an arbitrary model fine-tuned with LoRA approximates the target model. We theoretically prove that, under certain assumptions, LLMs are fine-tuned with LoRA approximate arbitrary classifiers. Theorem \ref{Lora:upperbound} supports the idea that LBC has a high generalization performance in tabular data classification.

\begin{theorem}
\textup{Let $f(\boldsymbol{x})$ represents the ReLU neural network to which LoRA is applied, with no activation function in the last layer, and $\Bar{f}(\boldsymbol{x})$ represents the target single-layer linear network. Let $g(x)$ is the logistic function $(1+e^{-x})^{-1}$. $\sigma(\boldsymbol{W})_i$ is the $i$-th greatest singular value of $\boldsymbol{W}$. $\boldsymbol{W}_l$ and $\overline{\boldsymbol{W}}$ are $l$-th layer weight matrix of the frozen model and the weight matrix of the target model, respectively.}
\textup{where $R_l, R_E$ are $Rank(W_l), Rank(\overline{\boldsymbol{W}}-\prod{\boldsymbol{W}_l})$, respectively. $L$ is the number of layers in $f$.}
\label{Lora:upperbound}
\end{theorem}

\subsection{Verbalizer}
Language generative models were adapted for classification tasks by utilizing verbalizers in the loss function. During the training process, using verbalizers encourages the model to generate semantically accurate responses rather than comparing labels precisely at the token level. Since this approach does not fit the model to fixed token-level labels, we can expect faster convergence when training generative language models for classification problems. LBC slightly modifies the structure of traditional LLMs in training and inference to use a verbalizer.

 Given a vector $\mathbf{Logit} = \{l_{w_1}, l_{w_2},  \ldots, l_{w_V}\}$, where $V$ is the vocabulary size and $l_{w_i}$ is the score for the word $w_i$ in the vocabulary, LBC's score for a single class $C_k$ is calculated as follows:
\begin{equation*}
\text{Score}(C_k) = \alpha_1l_{k} + \alpha_2\sum_{w \in S_k} l_w
\end{equation*}
where $k$ is the central word representing class $C_k$, $\alpha_1$ and $\alpha_2$ are the hyperparameters for the central word and synonyms, and $S_k$ is the set of synonyms of central word k. For example, if $k$ = 'Yes', then $S_k = \{\text{'yes'}, \text{'yeah'}, \text{'true'} ... \}$. The probability for $C_k$ is computed using a softmax function:
\begin{equation*}
P(C_k) = \frac{\exp({\text{Score}(C_k)})} {\sum_{k' \in K}{\exp(\text{Score}(C_{k'})})}
\end{equation*}
where $K$ is the set of central words of all classes. Besides, we modify the existing loss function as follows:
\begin{equation*}
J = \alpha_1\text{CE}(\mathbf{Logit}, L_{k}) + \alpha_2\sum_{w \in S_k} \text{CE}(\mathbf{Logit}, L_{w})
\end{equation*}

\section{Experiments}
\subsection{Experiment Settings}
We conducted experiments using reliable datasets that have been frequently used in studies, specifically selecting those that have been run multiple times on OpenML~\cite{OpenML2013}, Kaggle, or other benchmarks. Information about the eleven datasets is in Table \ref{data_info}. Details on the evaluation methods are in Appendix \ref{app:Eval}. As baselines, we selected five models, referred to as TMLs, which are known for their strong performance in tabular data classification. Details of the TMLs are in Appendix \ref{app:tml}. Additionally, to assess the performance improvements brought by LBC's three methodologies, we conducted direct comparisons with LIFT's methodology.

\subsection{OOV Setting}
To experiment with the performance of LBC on OOV tasks, it is essential to create scenarios where variables that do not exist in training appear in testing. However, we faced a problem because no existing tabular datasets fulfill this requirement. We randomly deleted 50\% of the variable columns in the original tabular dataset. As a result, variables that are deleted become OOV, not learned by the model during training, and emerge as new variables in the test. This allows for the assessment of LBC's ability to interpret OOVs. We compare the performance of TMLs and LBC with the data generated by this method.

\subsection{Avoiding Bias}
When fine-tuning LBC, if prompts consistently end with the same token, such as a question mark, the model may focus more on that token than on the actual variables when predicting class labels. This issue is particularly pronounced in datasets with class imbalance. To mitigate this, inserting random words at the end of the sentence helps reduce bias towards specific tokens. An example of the use of random word is shown in figure~\ref{fig:main_method}

\begin{table}[t!]
\centering
\parbox{0.8\columnwidth}{\caption{LBC vs TMLs in binary classification problems with 50\% randomly selected OOV situations. The models are trained with 50\% IVs, and LBCs add 50\% OOVs in the test prompts. LBC outperforms the TMLs on evaluation scores.}}
\label{tab:table_main_results_binaryclass}
\resizebox{0.8\columnwidth}{!}{%
\begin{tabular}{cccccccc}
\hline
\multicolumn{1}{l|}{Accuracy}      & DT      & KNN     & LogReg  & SVM      & \multicolumn{1}{c|}{XGBoost} & LBC - GPTJ      & LBC - Llama3                           \\ \hline
\multicolumn{1}{l|}{Blood}         & 72.67   & 69.33   & 75.33   & 75.33    & \multicolumn{1}{c|}{74.67}   & \textbf{76.00±0.00}      & \textbf{76.00±0.38}    \\
\multicolumn{1}{l|}{Breast Cancer} & 93.86   & 93.86   & 92.98   & 92.98    & \multicolumn{1}{c|}{92.98}   &94.15±1.01   & \textbf{94.44±0.50}    \\
\multicolumn{1}{l|}{Creditcard}    & 76.81   & 73.91   & 72.46   & 77.54    & \multicolumn{1}{c|}{76.09}   & \textbf{83.81±0.42}   & 80.84±0.54    \\
\multicolumn{1}{l|}{German}        & 71.00   & 71.50   & 77.50   & 71.50    & \multicolumn{1}{c|}{70.50}    & \textbf{78.50±0.86}  & 77.16±1.15   \\
\multicolumn{1}{l|}{ILPD}          & 70.94   & 60.68   & 72.65   & 70.94    & \multicolumn{1}{c|}{64.86}   & \textbf{75.05±0.84}   & 72.07±0.49   \\
\multicolumn{1}{l|}{Loan}          & 69.11   & 66.67   & 69.92   & 69.11    & \multicolumn{1}{c|}{59.35}   & 80.59±1.22   & \textbf{81.25±0.00}    \\
\multicolumn{1}{l|}{Salary}        & \textbf{85.00}   & 83.00   & 83.00      & 81.50     & \multicolumn{1}{c|}{83.00}      & 84.00±0.86    & 84.67±0.28 \\
\multicolumn{1}{l|}{Steel Plate}   & 80.21   & 79.69   & 73.78   & 78.15    & \multicolumn{1}{c|}{81.23}   &81.83±1.62   & \textbf{81.91±1.47}
    \\ \hline
\multicolumn{1}{l|}{Avg.}          & 77.53   & 74.83   & 77.18 & 75.01 & \multicolumn{1}{c|}{76.38}  & \textbf{81.74±0.85}   & 80.98±0.60    \\ \hline
                                   &         &         &         &          &                              &         &                                         \\ \hline
\multicolumn{1}{l|}{F1}            & DT      & KNN     & LogReg  & SVM      & \multicolumn{1}{c|}{XGBoost} & LBC - GPTJ     & LBC - Llama3                           \\ \hline
\multicolumn{1}{l|}{Blood}         & 0.68    & \textbf{0.73}    & 0.68    & 0.63     & \multicolumn{1}{c|}{\textbf{0.73}}    & 0.67±0.00   & 0.67±0.00     \\
\multicolumn{1}{l|}{Breast Cancer} & \textbf{0.94}    & \textbf{0.94}    & 0.93    & 0.93     & \multicolumn{1}{c|}{0.93}    & 0.93±0.00    & 0.93±0.00                               \\
\multicolumn{1}{l|}{Creditcard}    & 0.67    & 0.59    & 0.62    & 0.62     & \multicolumn{1}{c|}{0.67}    & \textbf{0.87±0.02}    & 0.81±0.01     \\
\multicolumn{1}{l|}{German}        & 0.73    & 0.77    & 0.77    & 0.73     & \multicolumn{1}{c|}{0.78}    & 0.71±0.01    & \textbf{0.78±0.01}                               \\
\multicolumn{1}{l|}{ILPD}          & \textbf{0.76}    & 0.71    & 0.73    & 0.74     & \multicolumn{1}{c|}{0.75}    & 0.75±0.00    & 0.75±0.00  \\
\multicolumn{1}{l|}{Loan}          & 0.70     & 0.70     & 0.71    & 0.70      & \multicolumn{1}{c|}{0.69}    & 0.76±0.01   & \textbf{0.78±0.01}     \\
\multicolumn{1}{l|}{Salary}        & 0.55    & 0.55    & 0.55    & 0.5      & \multicolumn{1}{c|}{\textbf{0.59}}    & 0.52±0.01    & 0.52±0.01                               \\
\multicolumn{1}{l|}{Steel Plate}   & 0.8     & 0.79    & 0.72    & 0.79     & \multicolumn{1}{c|}{\textbf{0.81}}    & 0.80±0.01    & 0.80±0.01   \\ \hline
\multicolumn{1}{l|}{Avg.}          & 0.72 & 0.71  & 0.70 & 0.68    & \multicolumn{1}{c|}{0.74} & 0.75±0.00   & \textbf{0.76±0.01}     \\ \hline
                                   &         &         &         &          &                              &         &                                         \\ \hline
\multicolumn{1}{l|}{AUC}           & DT      & KNN     & LogReg  & SVM      & \multicolumn{1}{c|}{XGBoost} & LBC - GPTJ     & LBC - Llama3                           \\ \hline
\multicolumn{1}{l|}{Blood}         & 0.67    & 0.61    & 0.67    & \textbf{0.68}     & \multicolumn{1}{c|}{\textbf{0.68}}    & 0.67±0.00     & 0.67±0.00    \\
\multicolumn{1}{l|}{Breast Cancer} & 0.97    & 0.98    & 0.98    & \textbf{0.99}     & \multicolumn{1}{c|}{\textbf{0.99}}    & \textbf{0.99±0.00}    & \textbf{0.99±0.00}  \\
\multicolumn{1}{l|}{Creditcard}    & 0.79    & 0.8     & 0.83    & 0.84     & \multicolumn{1}{c|}{0.80}     &\textbf{0.92±0.02}    & 0.85±0.01     \\
\multicolumn{1}{l|}{German}        & 0.67    & 0.69    & \textbf{0.80}     & 0.67     & \multicolumn{1}{c|}{0.69}    & \textbf{0.79±0.01}    & 0.78±0.01  \\
\multicolumn{1}{l|}{ILPD}          & 0.71    & 0.57    & 0.68    & 0.71     & \multicolumn{1}{c|}{0.71}    & \textbf{0.75±0.01}    & \textbf{0.75±0.00}     \\
\multicolumn{1}{l|}{Loan}          & 0.56    & 0.57    & 0.63    & 0.51     & \multicolumn{1}{c|}{0.53}    &\textbf{0.79±0.01}    &0.77±0.01     \\
\multicolumn{1}{l|}{Salary}        & 0.84    & 0.85    & 0.86    & 0.87     & \multicolumn{1}{c|}{0.86}    &\textbf{0.88±0.01}    & \textbf{0.88±0.01}     \\
\multicolumn{1}{l|}{Steel Plate}   & 0.87    & 0.89    & 0.89    & 0.89     & \multicolumn{1}{c|}{0.89}    &\textbf{0.90±0.00}    & \textbf{0.89±0.00}     \\ \hline
\multicolumn{1}{l|}{Avg.}          & 0.76  & 0.73 & 0.78   & 0.78   & \multicolumn{1}{c|}{0.78} &\textbf{0.84±0.00} & 0.82±0.00  \\ \hline
\end{tabular}
}

\end{table}

\begin{table}
\centering
\parbox{0.7\columnwidth}{\caption{LBC vs TMLs in multiclass classification problems with 50\% randomly selected OOV situations. LBC also outperforms the TMLs on evaluation scores in multiclass classification.}}
\label{tab:table_main_results_multiclass}
\resizebox{0.7\columnwidth}{!}{%
\begin{tabular}{lcccccc}
\hline
\multicolumn{1}{c|}{Accuracy}   & DT & KNN & LogReg & \multicolumn{1}{c|}{XGBoost} & LBC - GPTJ & LBC - Llama3 \\ \hline
\multicolumn{1}{c|}{CMC}        & 46.10     & 43.39    & 48.15    & \multicolumn{1}{c|}{45.42}   & 49.71±0.78 & \textbf{51.75±1.36}   \\
\multicolumn{1}{c|}{Restaurant} & 79.50     & 83.50    & 80.50    & \multicolumn{1}{c|}{84.50}   & 81.16±0.57 & \textbf{85.33±0.57}   \\
\multicolumn{1}{c|}{OGB}        & 50.00     & 51.50    & 55.00    & \multicolumn{1}{c|}{56.50}   & 56.73±1.44 & \textbf{62.51±0.79}   \\ \hline
\multicolumn{1}{c|}{Avg.}       & 58.53 & 59.46 & 61.22 & \multicolumn{1}{c|}{62.14}   & 62.53±0.93 & \textbf{66.53±0.91}   \\ \hline
                                &          &          &          &                              &            &              \\ \hline
\multicolumn{1}{c|}{F1}         & DT & KNN & LogReg & \multicolumn{1}{c|}{XGBoost} & LBC - GPTJ & LBC - Llama3 \\ \hline
\multicolumn{1}{c|}{CMC}        & 0.47     & 0.40     & 0.47     & \multicolumn{1}{c|}{0.45}    & 0.50±0.01  & \textbf{0.51±0.01}    \\
\multicolumn{1}{c|}{Restaurant} & 0.79     & 0.85     & 0.75     & \multicolumn{1}{c|}{0.85}    & 0.82±0.01  & \textbf{0.86±0.01}    \\
\multicolumn{1}{c|}{OGB}        & 0.34     & 0.51     & 0.54     & \multicolumn{1}{c|}{0.56}    & 0.54±0.01  & \textbf{0.57±0.02}    \\ \hline
\multicolumn{1}{c|}{Avg.}       & 0.53 & 0.59 & 0.59 & \multicolumn{1}{c|}{0.62}    & 0.62±0.01  & \textbf{0.64±0.01}   \\ \hline
\end{tabular}%
}
\end{table}

\section{Results}

\begin{wrapfigure}{r}{0.45\columnwidth}
\vspace{-15pt}
\centering
\includegraphics[width=0.45\columnwidth]{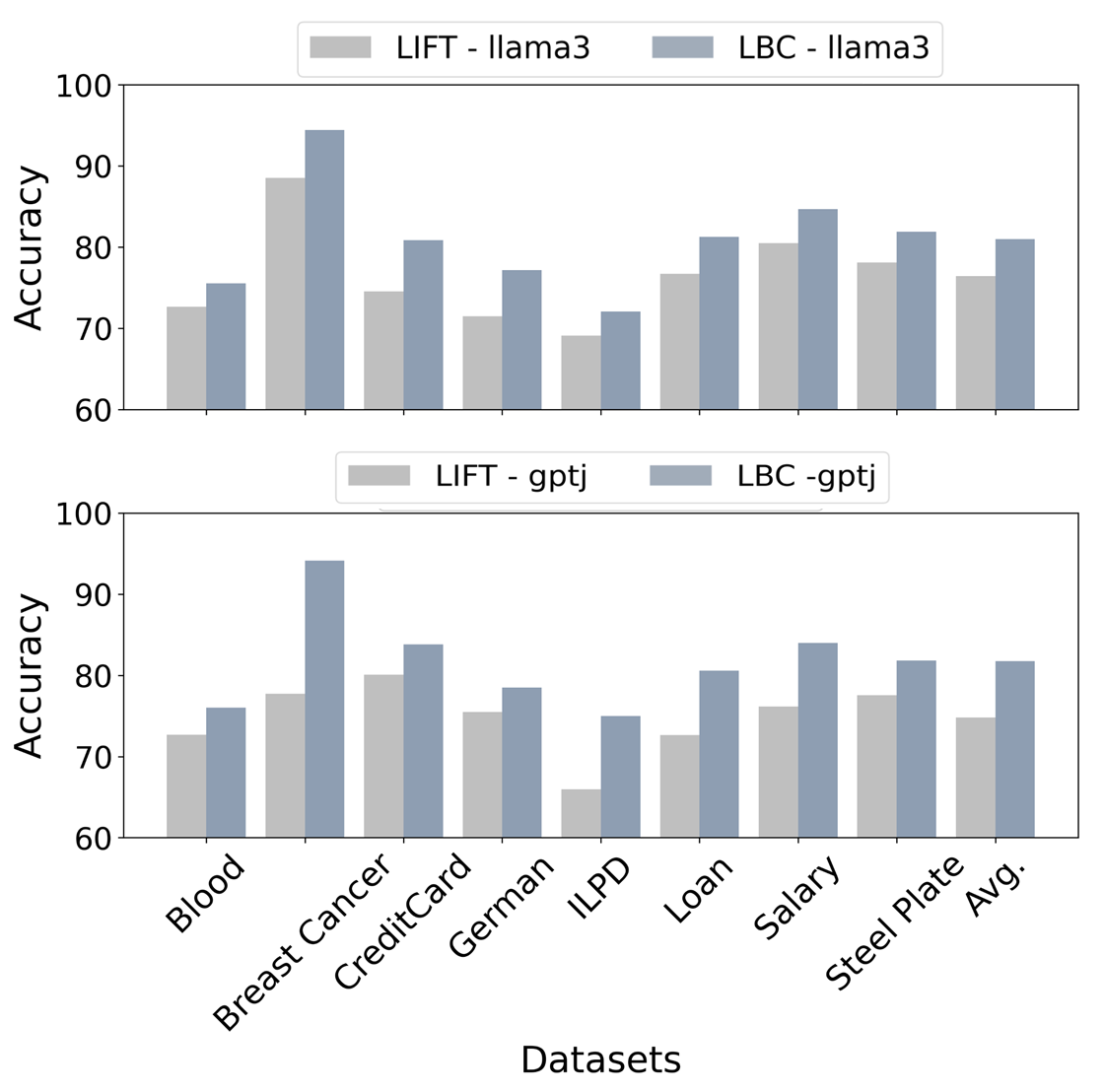}
\caption{LIFT vs LBC in 50\% randomly selected OOV situation. Both LLMs have a performance improvement when LBC's methodologies are applied rather than LIFT.}
\label{before_after_LBC_graph}
\end{wrapfigure}

\subsection{Performance in OOV tasks}
Table~\ref{tab:table_main_results_binaryclass} presents the accuracy, F1, and AUC scores of TMLs and LBCs on eight binary classification datasets after conducting 50\% OOV conversion. In the average rows for the evaluation metrics, LBC consistently outperforms the five TMLs in binary classification problems. Building on these results, we extended our experiments to multiclass classification tasks, as shown in Table~\ref{tab:table_main_results_multiclass}. LBCs continue to outperform TMLs, with LBC-Llama3, demonstrating strong performance in multiclass scenarios.

Table~\ref{tab:t-test} provides the statistical test results on the Accuracy scores from Table~\ref{tab:table_main_results_binaryclass} and Table~\ref{tab:table_main_results_multiclass}. For each dataset, a T-test was conducted between the model with the highest performance among LBCs and the best-performing TML. The null hypothesis, \( H_0: \text{Accuracy}_{\text{LBC-best}} = \text{Accuracy}_{\text{TML-best}} \), was rejected for seven out of eleven datasets, with a p-value less than 0.05 used as the criterion for rejection. This provides empirical evidence that LBC effectively utilizes pre-trained knowledge to make accurate interpretations in OOV situations. Further analysis of this capability is discussed in Section~\ref{tab:Zeroshot_LBC_ability}.

\begin{wraptable}{r}{0.5\columnwidth}
\vspace{-15pt}
\caption{Accuracy evaluation of the proposed models. We perform five repeated experiments on the model with the highest performance among the TML and LBC methods, and conduct a t-test. The left two columns represent the mean accuracy of the repeated experiments. The p-values less than 0.05 are highlighted in bold and marked with an asterisk (*). For seven of the eleven datasets, it is valid that LBC outperforms TML.}
\centering
\resizebox{0.5\columnwidth}{!}{%
\begin{tabular}{l|cc|cc}
\hline
Datasets      & LBC-Best   & TML-Best   & T-stats & P-value \\ \hline
Blood         & 76.00 & 75.55 & 1.71    & 0.12    \\
Breast Cancer & 94.44 & 93.27 & 2.94    & \textbf{0.01*} \\
Creditcard    & 83.81 & 77.37 & 6.24    & \textbf{0.00*} \\
German        & 78.50 & 76.66 & 3.20    & \textbf{0.03*} \\
ILPD          & 75.05 & 72.19 & 3.43    & \textbf{0.02*} \\
Loan          & 81.25 & 70.27 & 18.72   & \textbf{0.00*} \\
Salary        & 84.67 & 84.83 & -0.90   & 0.39    \\
Steel Plate   & 81.91 & 80.51 & 1.89    & 0.09    \\
CMC           & 51.75 & 47.88 & 6.11    & \textbf{0.00*} \\
OGB           & 62.51 & 57.16 & 3.03    & \textbf{0.03*} \\ 
Restaurant    & 85.33 & 84.66 & 0.60    & 0.57    \\ \hline
\end{tabular}
}
\label{tab:t-test}
\end{wraptable}

Figure~\ref{before_after_LBC_graph} shows how much LBC's methodologies improve the performance of LLM on the OOV task. There is a significant difference in performance between using LBC and using LIFT, which does not incorporate LBC's three methodologies. This demonstrates that, in addition to the advantage of LLM's pre-trained knowledge in interpreting OOVs, LBC's methodologies have a clear and positive impact on performance.

\begin{figure}
\centering
\includegraphics[width = \textwidth]{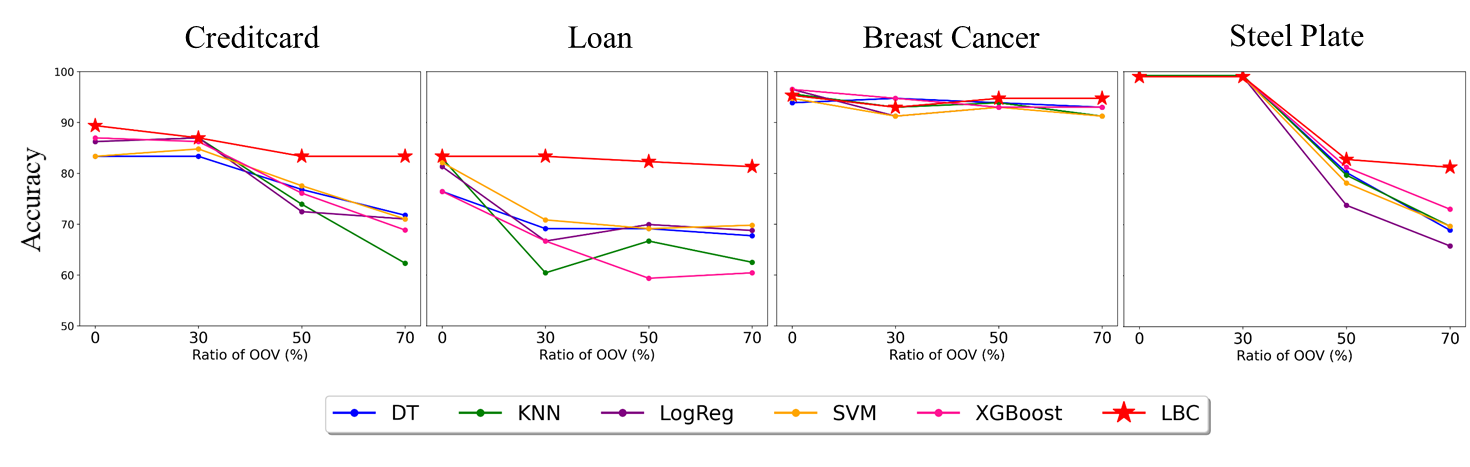}
\caption{Graph of accuracy changing over OOV ratio (\%): We observed the accuracy change of TMLs and LBCs by increasing the OOV ratio from 0, 30, 50, and 70 (\%) for four datasets. Comparing the accuracy reduction of TMLs and LBCs, the reduction of LBCs is smaller compared to TMLs. It demonstrates that LBCs interpret OOVs, unlike TMLs.}
\label{oov_ratio_graph}
\end{figure}

To validate the ability of LBC to perform well on OOV tasks, we conduct experiments on four datasets with different OOV ratios. In each dataset, we vary the OOV ratio to 0\%, 30\%, 50\%, and 70\% and observe the model's accuracy change. Figure~\ref{oov_ratio_graph} shows that for TMLs, the performance decreases significantly as the OOV ratio increases. In contrast, LBC shows no decrease in accuracy as the OOV ratio increases or the decrease is small compared to TMLs. These findings suggest that LBC can effectively utilize the pre-trained knowledge of LLMs to outperform traditional ML methods even as the percentage of OOVs increases. 
\label{sec:6.2}

\subsection{LBC's Ability to utilize Pre-Trained Knowledge} \label{tab:Zeroshot_LBC_ability}
In this section, we specifically investigate how LBC uses their pre-trained knowledge to interpret OOVs in the OOV tasks. We conduct an experiment to observe the pre-trained bias for variables using an LBC that has been trained with the structure of data without any information about the variables. For datasets with a "Yes" or "No" answer, The structure of the data is as follows:

\small{
\begin{align*}
\text{Prompt} &= \text{string(Start of sentence)} + \text{string('Variable name' is [Variable value])} + \text{string(Question)} \\ \text{Answer} &= \text{Yes@@@ or No@@@}
\end{align*}
}

\normalsize
\noindent In the training process, the prompt structure is utilized as it is, with experimental adjustments made to balance the likelihood of the trained LBC predicting 'Yes' or 'No'. During testing, we replace the names and values of various variables in the 'variable name' and 'variable value' placeholders within the prompts to evaluate the pre-trained biases of LBC towards those variables.

\begin{figure}
\centering
\includegraphics[width = 0.6\columnwidth]{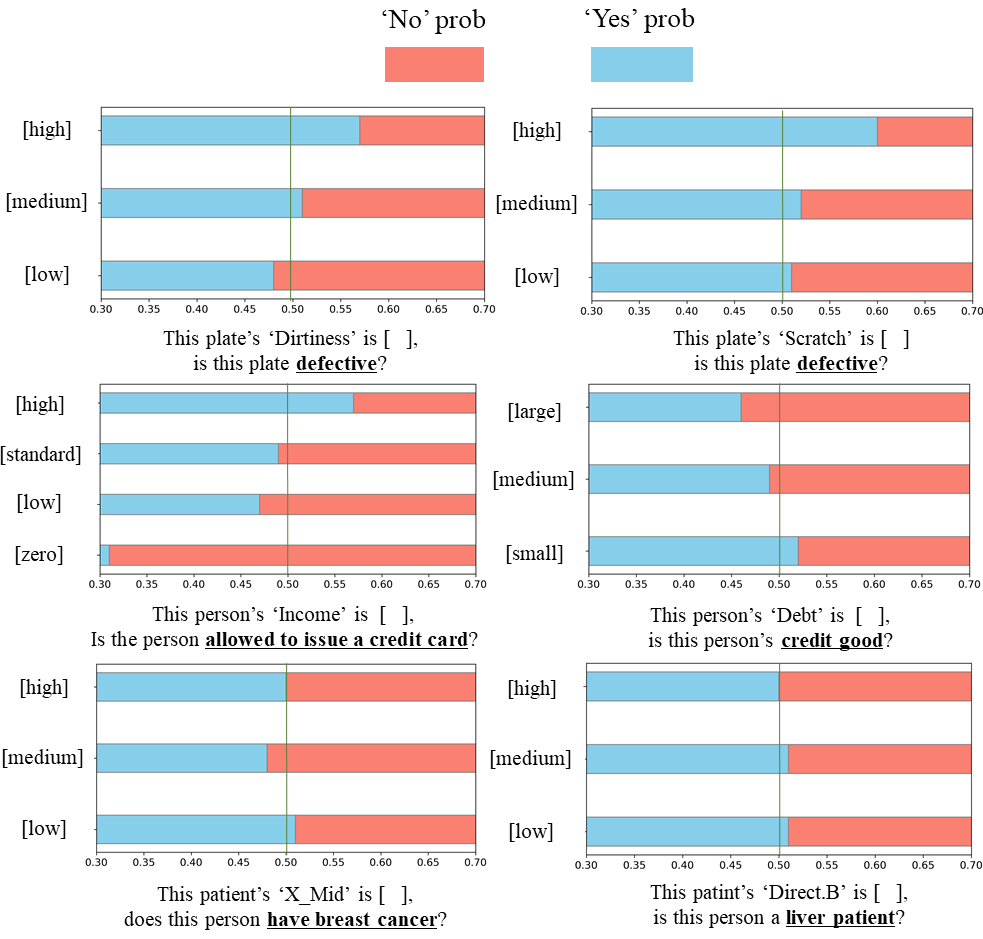}
\caption{Observing how LBC applies its pre-trained knowledge to prompts about OOVs, thereby revealing biases in its pre-trained knowledge. Intuitively, LBC has a bias toward making its predictions closer to the correct answer. However, it is not responsive to special variable names that do not have a word meaning.}
\label{zeroshot_vias_graph}
\end{figure}

Figure~\ref{zeroshot_vias_graph} presents the outcomes for several variables of high importance in this experiment. It is evident that LBC leverages pre-trained knowledge to approximate the probabilities for variables not learned during training closer to the correct answers. Notably, the interpretation that the graph for "income" shows a high risk of issuing a credit card to an individual with an income of "0" matches with the actual distribution in the Creditcard dataset. However, for the unique variables in the table dataset, such as "Direct.B", which does not have the meaning of a common word, LBC shows almost balanced results and tends to make predictions without any clear bias. This shows that LBC maintains a neutral approach to uninterpretable variables, and maintains an even probability distribution without any particular tendency. These results support the high performance of LBC in handling OOV tasks.

\begin{wrapfigure}{r}{0.6\columnwidth}
\centering
\includegraphics[width=0.6\columnwidth]{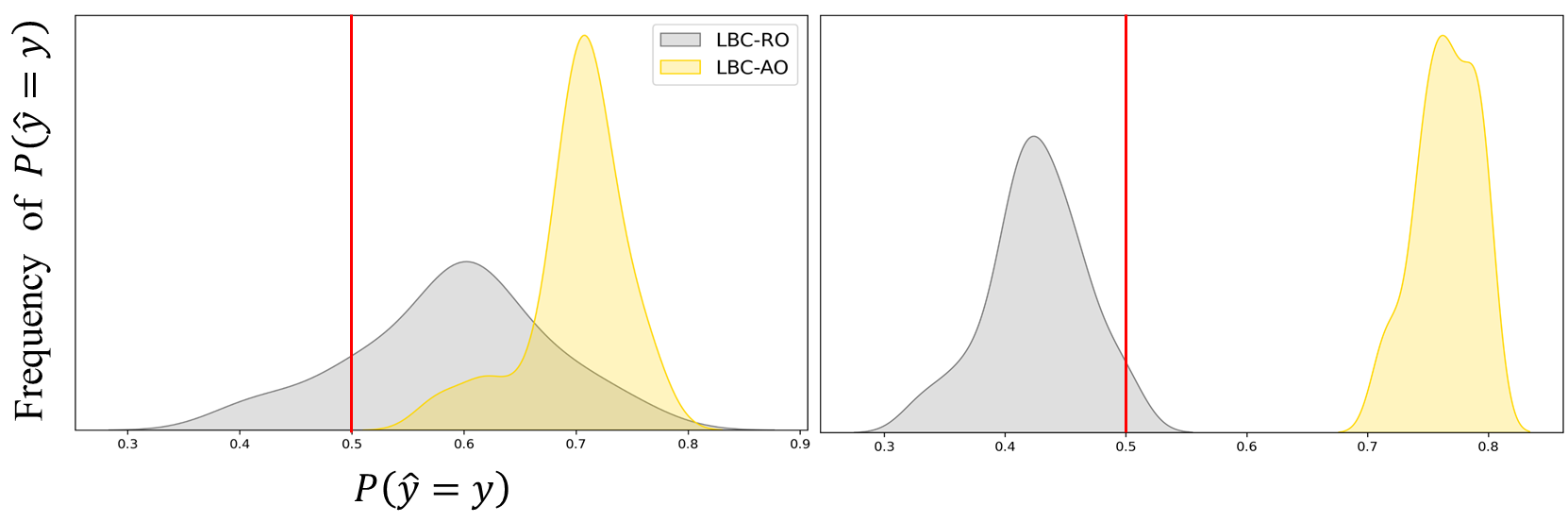}
\caption{Frequeny of $P(\hat{y} = y)$ for the two prompt generation methods. We repeated the prompt generation 100 times for each of the two randomly selected examples from the Creditcard dataset in two ways: random order (RO) and advanced order (AO). The horizontal axis represents the model's probability for the correct class $y$, and the vertical axis shows frequency. The AO method provides more consistent and accurate results than RO. The red vertical line indicates the prediction boundary, where the differences between the two methods lead to varied predictions.}
\label{prob_distribution}
\end{wrapfigure}

\vspace{1cm}

\subsection{Importance of Advanced Prompt}
In this section, we examine how Advanced Prompts, such as "Consider the order of variables" or "Add an indicator," used to generate test prompts, affect LBC's probability output.

To verify the importance of the variables' order, we experiment with repeatedly generating two types of prompts by randomly selecting an instance from the tabular data:  One, where the order of all variables is randomized (LBC-RO), and the other, where the order of the IVs matches to the IV of the training data, and only the order of the OOVs are randomized (LBC-AO). We randomly select two instances from the Creditcard dataset and generate 100 different prompts for each instance with the RO and AO methods, respectively, to compare the probability distributions generated by LBC for the two methods. Figure~\ref{prob_distribution} illustrates the performance difference between prompts where the order of variables is matched with the training data and those where it is not. LBC-RO exhibits a large variance in the probability distribution, leading to variations in the model's predictions for a single data instance. In contrast, LBC-AO shows a small variance in the probability distribution, which means that the model makes consistent predictions.

\begin{wrapfigure}{r}{0.6\columnwidth}
\centering
\includegraphics[width=0.6\columnwidth]{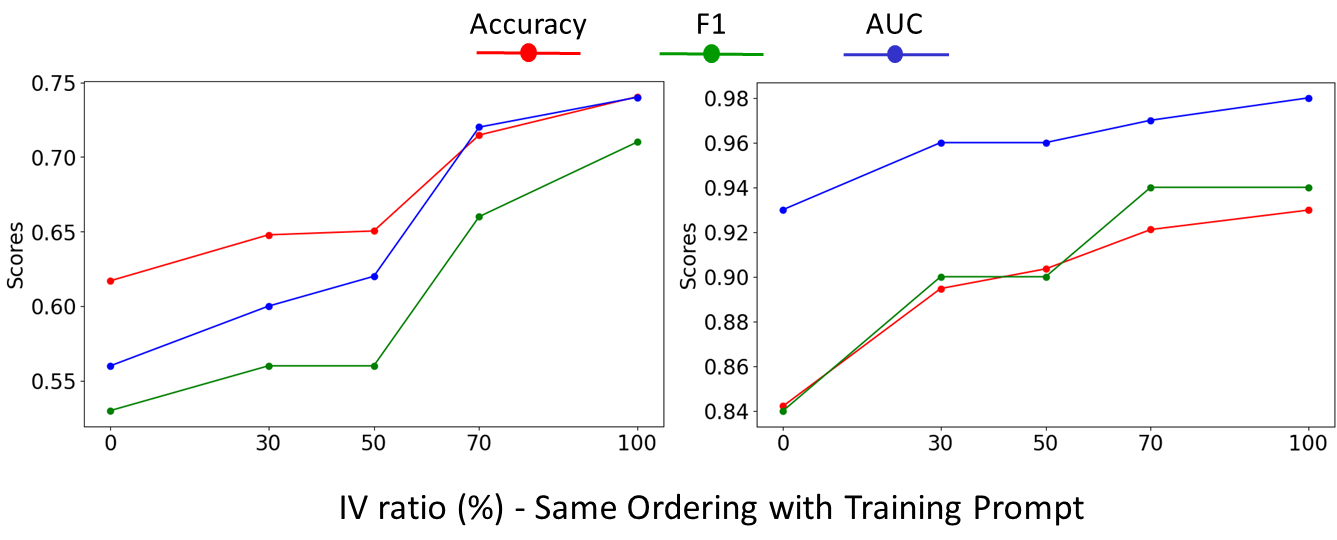}
\caption{The changes in scores according to the ratio of IVs in the test prompt that maintains the same order as the IVs in the training prompt. Both the training and test prompts consist of only the IVs used in table~\ref{tab:table_main_results_binaryclass}. As the ratio of the same IV order increases between the training and test prompts, all scores improve. This demonstrates the importance of applying the same order of IVs in the test prompt as in the training prompt.}
\label{Fix_IV_ratio_graph}
\end{wrapfigure}

To further investigate the benefits of matching the order of IVs of test prompts with the training prompts, we compose the training and test data using only the IVs, excluding the OOVs selected from the Steel Plate dataset used in Table~\ref{tab:table_main_results_binaryclass}. Then, for the variables that make up the test prompt, we experiment with increasing the ratio of variables in the same order as the variable order of the training prompt to check the scores for the three evaluation metrics. Figure~\ref{Fix_IV_ratio_graph} illustrates the scores for the three evaluation metrics. As the IVs ratio increases, the performance improves on all three metrics. This shows that LBC performs best when the test data follows the same variable order as the training data.
\label{order experiment} 

\vspace{-0.4cm}

\subsection{LBC - Black-box LLM}
Although it is possible to configure LBC using the latest LLM, most of the latest models are black-box, so we conduct in-context learning experiments. The model with the LBC methodology incorporates the categorical change, advanced order, and indicator methodologies. Table~\ref{tab:icl} compares the Accuracy, F1, and AUC scores before and after applying the LBC methodology. We use GPT-3.5 as the model, and performance improves significantly on all datasets when we add the LBC methodology. This demonstrates that the LBC methodology can also be applied to black-box LLMs to improve performance. The performance on its own is not high because of a small number of training examples for in-context learning compared to fine-tuning, and it is considered to be better when it is fine-tuned.

\begin{wraptable}{r}{0.5\columnwidth}
\vspace{-15pt}
\caption{Comparison of three performance metrics before and after applying LBC's methodology using in-context learning to a modern black-box LLM}
\label{tab:icl}
\resizebox{0.5\columnwidth}{!}{%
\begin{tabular}{cccccc}
\hline
\multicolumn{1}{c|}{Accuracy}    & Creditcard     & German         & ILPD           & \multicolumn{1}{c|}{Loan}           & Avg.\\ \hline
\multicolumn{1}{c|}{GPT3.5}      & 60.15          & 63.50          & 62.75          &\multicolumn{1}{c|}{63.54}         & 62.50\\
\multicolumn{1}{c|}{LBC-GPT3.5} & \textbf{69.57} & \textbf{67.50} & \textbf{63.25} & \multicolumn{1}{c|}{\textbf{66.67}} & \textbf{66.74}\\ \hline
                                 &                &                &                &               & \\ \hline
\multicolumn{1}{c|}{F1}          & Creditcard     & German         & ILPD           & \multicolumn{1}{c|}{Loan}          & Avg.\\ \hline
\multicolumn{1}{c|}{GPT3.5}      & 0.61           & 0.55           & 0.57           & \multicolumn{1}{c|}{0.60}          & 0.58 \\
\multicolumn{1}{c|}{LBC-GPT3.5} & \textbf{0.69}  & \textbf{0.57}  & \textbf{0.61}  & \multicolumn{1}{c|}{\textbf{0.66}} & \textbf{0.63}\\ \hline
                                 &                &                &                &               & \\ \hline
\multicolumn{1}{c|}{AUC}         & Creditcard     & German         & ILPD           & \multicolumn{1}{c|}{Loan}          & Avg.\\ \hline
\multicolumn{1}{c|}{GPT3.5}      & 0.60           & 0.52           & 0.51           & \multicolumn{1}{c|}{0.55}          & 0.54\\
\multicolumn{1}{c|}{LBC-GPT3.5} & \textbf{0.69}  & \textbf{0.57}  & \textbf{0.54}  & \multicolumn{1}{c|}{\textbf{0.59}} &  \textbf{0.60}\\ \hline
\end{tabular}
}
\end{wraptable}

\section{Conclusion}
In this work, we propose LBC to solve OOV tasks.  LBCs utilize prompt-based inference, which allows information about OOVs to be added to prompts in a straightforward way and enables understanding of the new information through pre-trained knowledge. LBC's three methodologies maximize the above advantages to achieve high performance on OOV tasks. As a result, utilizing LLMs' pre-trained knowledge is a key strategy for solving the OOV task, and we plan to combine it with various statistical methods. LBC is the first approach to apply pre-trained LLM to OOV tasks.

\section{Limitations}
Based on our three methodologies, LBC demonstrates superior performance over TML in addressing the OOV generalization problem, leveraging pre-trained knowledge and the contextual understanding capabilities of LLMs. However, several limitations still exist. The first limitation is the potential presence of data that requires knowledge not covered in pre-training. When column names are unintelligible or involve extremely recent information not included in pre-training, LBC faces difficulties in interpretation. The second limitation is that LBC requires more resources compared to TML. In terms of training time and GPU specifications, LBC demands higher costs than TML. Therefore, in cases where the information content of OOV is low or when the problem does not involve OOV, LBC is less suitable compared to TML.

\clearpage

\bibliography{arxiv_reference}

\begin{thebibliography}{28}
\providecommand{\natexlab}[1]{#1}
\providecommand{\url}[1]{\texttt{#1}}
\expandafter\ifx\csname urlstyle\endcsname\relax
  \providecommand{\doi}[1]{doi: #1}\else
  \providecommand{\doi}{doi: \begingroup \urlstyle{rm}\Url}\fi

\bibitem[Brown et~al.(2020)Brown, Mann, Ryder, Subbiah, Kaplan, Dhariwal, Neelakantan, Shyam, Sastry, Askell, et~al.]{gpt3}
Tom Brown, Benjamin Mann, Nick Ryder, Melanie Subbiah, Jared~D Kaplan, Prafulla Dhariwal, Arvind Neelakantan, Pranav Shyam, Girish Sastry, Amanda Askell, et~al.
\newblock Language models are few-shot learners.
\newblock \emph{Advances in neural information processing systems}, 33:\penalty0 1877--1901, 2020.

\bibitem[Buscema et~al.(2010)Buscema, Terzi, and Tastle]{misc_steel_plates_faults_198}
Massimo Buscema, Stefano Terzi, and William Tastle.
\newblock Steel plates faults.
\newblock UCI Machine Learning Repository, 2010.
\newblock {DOI}: https://doi.org/10.24432/C5J88N.

\bibitem[Chen and Guestrin(2016)]{xgboost}
Tianqi Chen and Carlos Guestrin.
\newblock Xgboost: A scalable tree boosting system.
\newblock In \emph{Proceedings of the 22nd acm sigkdd international conference on knowledge discovery and data mining}, pages 785--794, 2016.

\bibitem[Devlin et~al.(2018)Devlin, Chang, Lee, and Toutanova]{bert}
Jacob Devlin, Ming-Wei Chang, Kenton Lee, and Kristina Toutanova.
\newblock Bert: Pre-training of deep bidirectional transformers for language understanding.
\newblock \emph{arXiv preprint arXiv:1810.04805}, 2018.

\bibitem[Dinh et~al.(2022)Dinh, Zeng, Zhang, Lin, Gira, Rajput, Sohn, Papailiopoulos, and Lee]{lift}
Tuan Dinh, Yuchen Zeng, Ruisu Zhang, Ziqian Lin, Michael Gira, Shashank Rajput, Jy-yong Sohn, Dimitris Papailiopoulos, and Kangwook Lee.
\newblock Lift: Language-interfaced fine-tuning for non-language machine learning tasks.
\newblock \emph{Advances in Neural Information Processing Systems}, 35:\penalty0 11763--11784, 2022.

\bibitem[Dreher et~al.(2023)Dreher, Ayala, Schellenberg, Hübner, Nölke, Adler, Seidlitz, Sellner, Studier-Fischer, Gröhl, Nickel, Köthe, Seitel, and Maier-Hein]{Dreher_2023}
Kris~K. Dreher, Leonardo Ayala, Melanie Schellenberg, Marco Hübner, Jan-Hinrich Nölke, Tim~J. Adler, Silvia Seidlitz, Jan Sellner, Alexander Studier-Fischer, Janek Gröhl, Felix Nickel, Ullrich Köthe, Alexander Seitel, and Lena Maier-Hein.
\newblock \emph{Unsupervised Domain Transfer with Conditional Invertible Neural Networks}, page 770–780.
\newblock Springer Nature Switzerland, 2023.
\newblock ISBN 9783031439070.
\newblock \doi{10.1007/978-3-031-43907-0_73}.
\newblock URL \url{http://dx.doi.org/10.1007/978-3-031-43907-0_73}.

\bibitem[Guo et~al.(2023)Guo, Wildberger, and Sch{\"o}lkopf]{oov-gen}
Siyuan Guo, Jonas Wildberger, and Bernhard Sch{\"o}lkopf.
\newblock Out-of-variable generalization.
\newblock \emph{arXiv preprint arXiv:2304.07896}, 2023.

\bibitem[Guo et~al.(2024)Guo, Wildberger, and Schölkopf]{guo2024outofvariable}
Siyuan Guo, Jonas Wildberger, and Bernhard Schölkopf.
\newblock Out-of-variable generalization for discriminative models, 2024.

\bibitem[Hofmann(1994)]{credit-g}
Hans Hofmann.
\newblock {Statlog (German Credit Data)}.
\newblock UCI Machine Learning Repository, 1994.
\newblock {DOI}: https://doi.org/10.24432/C5NC77.

\bibitem[Hollmann et~al.(2022)Hollmann, M{\"u}ller, Eggensperger, and Hutter]{tabpfn}
Noah Hollmann, Samuel M{\"u}ller, Katharina Eggensperger, and Frank Hutter.
\newblock Tabpfn: A transformer that solves small tabular classification problems in a second.
\newblock \emph{arXiv preprint arXiv:2207.01848}, 2022.

\bibitem[Hu et~al.(2021)Hu, Shen, Wallis, Allen-Zhu, Li, Wang, Wang, and Chen]{lora}
Edward~J Hu, Yelong Shen, Phillip Wallis, Zeyuan Allen-Zhu, Yuanzhi Li, Shean Wang, Lu~Wang, and Weizhu Chen.
\newblock Lora: Low-rank adaptation of large language models.
\newblock \emph{arXiv preprint arXiv:2106.09685}, 2021.

\bibitem[Hu et~al.(2022)Hu, Ding, Wang, Liu, Wang, Li, Wu, and Sun]{verbalizer2}
Shengding Hu, Ning Ding, Huadong Wang, Zhiyuan Liu, Jingang Wang, Juanzi Li, Wei Wu, and Maosong Sun.
\newblock Knowledgeable prompt-tuning: Incorporating knowledge into prompt verbalizer for text classification, 2022.

\bibitem[Kharoua(2024{\natexlab{a}})]{rabie_el_kharoua_2024_OGB}
Rabie~El Kharoua.
\newblock Predict online gaming behavior dataset, 2024{\natexlab{a}}.
\newblock URL \url{https://www.kaggle.com/dsv/8742674}.

\bibitem[Kharoua(2024{\natexlab{b}})]{rabie_el_kharoua_2024_restaurant}
Rabie~El Kharoua.
\newblock Predict restaurant menu items profitability, 2024{\natexlab{b}}.
\newblock URL \url{https://www.kaggle.com/dsv/8743558}.

\bibitem[Kohavi(1996)]{misc_census_income_20}
Ron Kohavi.
\newblock {Census Income}.
\newblock UCI Machine Learning Repository, 1996.
\newblock {DOI}: https://doi.org/10.24432/C5GP7S.

\bibitem[Lim(1997)]{misc_contraceptive_method_choice_30}
Tjen-Sien Lim.
\newblock {Contraceptive Method Choice}.
\newblock UCI Machine Learning Repository, 1997.
\newblock {DOI}: https://doi.org/10.24432/C59W2D.

\bibitem[Mirza(2023)]{mazaharul_hasnine_mirza_2023}
Mazaharul~Hasnine Mirza.
\newblock Loan data set, 2023.
\newblock URL \url{https://www.kaggle.com/dsv/5149638}.

\bibitem[Quinlan()]{misc_credit_approval_27}
J.~R. Quinlan.
\newblock {Credit Approval}.
\newblock UCI Machine Learning Repository.
\newblock {DOI}: https://doi.org/10.24432/C5FS30.

\bibitem[Radford et~al.(2018)Radford, Narasimhan, Salimans, Sutskever, et~al.]{gpt1}
Alec Radford, Karthik Narasimhan, Tim Salimans, Ilya Sutskever, et~al.
\newblock Improving language understanding by generative pre-training.
\newblock 2018.

\bibitem[Ramana and Venkateswarlu(2012)]{indian_liver_patient_dataset}
Bendi Ramana and N.~Venkateswarlu.
\newblock {ILPD (Indian Liver Patient Dataset)}.
\newblock UCI Machine Learning Repository, 2012.
\newblock {DOI}: https://doi.org/10.24432/C5D02C.

\bibitem[Schick and Schütze(2021)]{verbalizer}
Timo Schick and Hinrich Schütze.
\newblock It's not just size that matters: Small language models are also few-shot learners, 2021.

\bibitem[Tzeng et~al.(2015)Tzeng, Hoffman, Darrell, and Saenko]{DBLP:journals/corr/TzengHDS15}
Eric Tzeng, Judy Hoffman, Trevor Darrell, and Kate Saenko.
\newblock Simultaneous deep transfer across domains and tasks.
\newblock \emph{CoRR}, abs/1510.02192, 2015.
\newblock URL \url{http://arxiv.org/abs/1510.02192}.

\bibitem[Vanschoren et~al.(2013)Vanschoren, van Rijn, Bischl, and Torgo]{OpenML2013}
Joaquin Vanschoren, Jan~N. van Rijn, Bernd Bischl, and Luis Torgo.
\newblock Openml: Networked science in machine learning.
\newblock \emph{SIGKDD Explorations}, 15\penalty0 (2):\penalty0 49--60, 2013.
\newblock \doi{10.1145/2641190.2641198}.
\newblock URL \url{http://doi.acm.org/10.1145/2641190.2641198}.

\bibitem[Wang and Komatsuzaki(2021)]{gpt-j}
Ben Wang and Aran Komatsuzaki.
\newblock {GPT-J-6B: A 6 Billion Parameter Autoregressive Language Model}.
\newblock \url{https://github.com/kingoflolz/mesh-transformer-jax}, May 2021.

\bibitem[Yan et~al.(2024)Yan, Zheng, Xu, Zhu, Chen, Sun, Wu, and Chen]{yan2024makingpretrainedlanguagemodels}
Jiahuan Yan, Bo~Zheng, Hongxia Xu, Yiheng Zhu, Danny~Z. Chen, Jimeng Sun, Jian Wu, and Jintai Chen.
\newblock Making pre-trained language models great on tabular prediction, 2024.
\newblock URL \url{https://arxiv.org/abs/2403.01841}.

\bibitem[Yeh(2008)]{blood_transfusion_service_center}
I-Cheng Yeh.
\newblock {Blood Transfusion Service Center}.
\newblock UCI Machine Learning Repository, 2008.
\newblock {DOI}: https://doi.org/10.24432/C5GS39.

\bibitem[Zeng and Lee(2023)]{expressive-power-of-lora}
Yuchen Zeng and Kangwook Lee.
\newblock The expressive power of low-rank adaptation.
\newblock \emph{arXiv preprint arXiv:2310.17513}, 2023.

\bibitem[Zwitter and Soklic(1988)]{misc_breast_cancer_14}
Matjaz Zwitter and Milan Soklic.
\newblock {Breast Cancer}.
\newblock UCI Machine Learning Repository, 1988.
\newblock {DOI}: https://doi.org/10.24432/C51P4M.

\end{thebibliography}

\appendix
\section{Hyperparameters for Experiments}
The hyperparameters for our experiments were set as follows: Learning rate in \{1e-3, 1e-4, 1e-5\}, LoRA rank in \{8, 16, 48, 144, 196\}, Epoch in \{5, 7, 10, 12\}. We conduct the grid search over those hyperparameters. Verbalizer $\alpha_1$ in \{0.6, 0.7, 0.8, 0.9\}
\section{Proof of Theorem \ref{Lora:upperbound}}
According to~\cite{expressive-power-of-lora}, an arbitrary model fine-tuned with LoRA approximates the target model. We extend this theory and theoretically prove that, under certain assumptions, LLMs are fine-tuned with LoRA approximate arbitrary classifiers. Theorem \ref{Lora:upperbound} supports the idea that LBC has a high generalization performance in tabular data classification.

\begin{figure*}
\centering
\includegraphics[width = \textwidth]{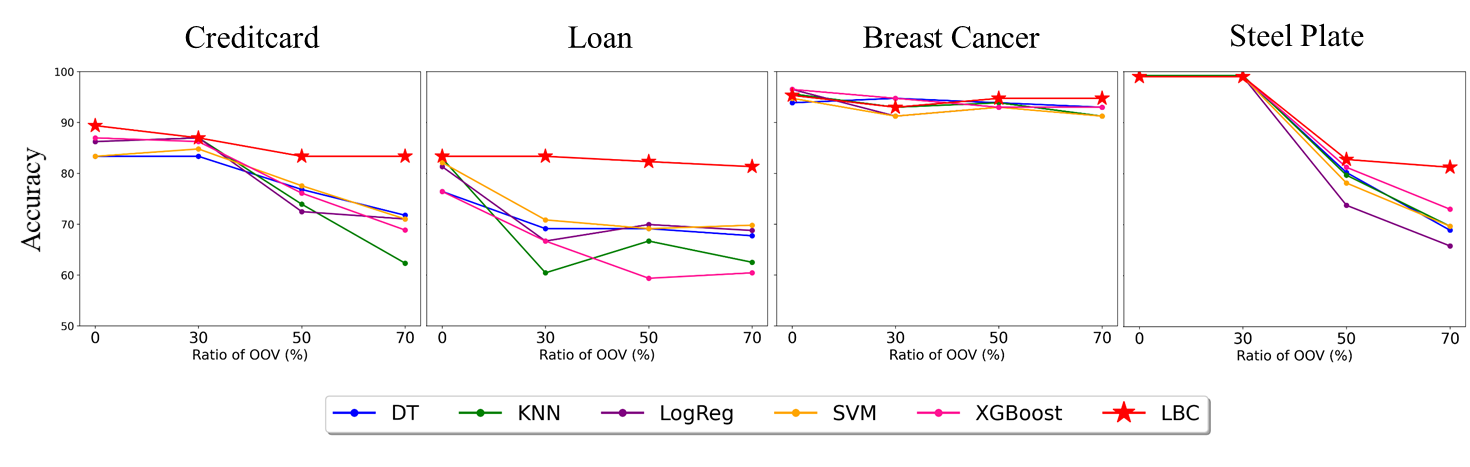}
\caption{Graph of accuracy changing over OOV ratio (\%): We observed the accuracy change of TMLs and LBCs by increasing the OOV ratio from 0, 30, 50, and 70 (\%) for four datasets. Comparing the accuracy reduction of TMLs and LBCs, the reduction of LBCs is smaller compared to TMLs. It demonstrates that LBCs interpret OOVs, unlike TMLs.}
\end{figure*}

\begin{lemma}
The logistic function $g(x)=(1+e^{-x})^{-1}$ is Lipschitz continuous with a Lipschitz constant of $1/4$.
\label{lemma1}
\end{lemma}

Proof of Lemma \ref{lemma1} A function $f$\,:\,$\mathbb{R}\to\mathbb{R}$ is Lipschitz continuous if

\begin{equation}
    \exists{K>0},\forall{x_1,x_2\in\mathbb{R}},|f(x_1)-f(x_2)|\le K|x_1-x_2|.
\end{equation}

By substituting $f$ with $g$, and considering that $g$ is
a monotonic function, we can obtain the following expression:
\begin{align*}
    &\frac{g(x_1)-g(x_2)}{x_1-x_2}\le K.
\end{align*}
By the mean value theorem,
\begin{align*}
    &g'(c)=\frac{g(x_2)-g(x_1)}{x_2-x_1}\le K,\text{and}\\
    &0< g'(c)\le\frac{1}{4}\\
    &\tag*{(\text{$g'(c)=g(c)(1-g(c)) \wedge 0<g(c)<1$})}\\
    &\to K\ge\frac{1}{4}.
\end{align*}
A new theorem, which is a variation of Lemma 11 of ~\cite{expressive-power-of-lora}, can be proposed using Lemma 1 above.\\
\ \\
\textbf{Theorem 1.}
\textup{Let $f(\boldsymbol{x})$ represents the ReLU neural network to which LoRA is applied, with no activation function in the last layer, and $\Bar{f}(\boldsymbol{x})$ represents the target single-layer linear network. Let $g(x)$ is the logistic function $(1+e^{-x})^{-1}$. $\sigma(\boldsymbol{W})_i$ is the $i$-th greatest singular value of $\boldsymbol{W}$. $\boldsymbol{W}_l$ and $\overline{\boldsymbol{W}}$ are $l$-th layer weight matrix of the frozen model and the weight matrix of the target model, respectively.}

\begin{align*}
    &\mathbb{E}\left\|g(f(\boldsymbol{x}))-g(\Bar{f}(\boldsymbol{x}))\right\|_2^2\\
    &\le\frac{1}{16}\mathbb{E}\left\|(f(\boldsymbol{x})-\Bar{f}(\boldsymbol{x}))\right\|_2^2\\
    &\tag*{(\text{g is $1/4$ Lipschitz by Lemma 1})}\\
    &\le\frac{1}{16}\|\mathbb{E}(\boldsymbol{xx}^T)\|_\text{F}\,\sigma^2\!\left(\overline{\boldsymbol{W}}-\prod\!\boldsymbol{W}_l\right)_{\min(\sum_{l=1}^LR_l, R_E)+1}.
\end{align*}

\textup{where $R_l, R_E$ are $Rank(W_l), Rank(\overline{\boldsymbol{W}}-\prod{\boldsymbol{W}_l})$, respectively. $L$ is the number of layers in $f$.}

\section{Traditional Machine Learning Models} \label{app:tml}
For Traditional Machine Learning Models, we selected 5 models. For tree-based models, we chose Decision Tree and XGBoost. Tree-based models have strong performance in tabular data classification. We also included K-Nearest Neighbor, Logistic Regression, and Support Vector Machine to increase the diversity of the models.

\textbf{Decision Tree.}
A Decision Tree (DT) is a model used for classification and regression tasks. The model trains on data to make predictions based on simple decision rules. The advantage of decision trees is that they capture non-linear patterns in data, and the results of the model are easy to interpret.

\textbf{K-Nearest Neighbor}
The K-Nearest Neighbor (KNN) algorithm is used in classification and regression to make predictions based on the data of the K closest neighbors.

\textbf{Logistic Regression}
Logistic regression (LogReg) is a model often used for classification problems. This model is often used when the outcome is binary, and it estimates probabilities to perform classification based on decision boundaries.

\textbf{Support Vector Machine}
A Support Vector Machine (SVM) is a machine learning model used for classification and regression problems. The model finds the optimal decision boundary to divide a given set of data into categories, and in this study, we used a Radial Basis Function (RBF) kernel.

\textbf{XGBoost} XGBoost is a high-performance machine learning model based on the Gradient Boosting algorithm, which is a decision tree-based ensemble learning method that combines multiple tree models. At each step, XGBoost adds a new model to reduce the error of the previous model and uses the Gradient Boosting technique in the process. XGBoost is a model that is state-of-the-art on many benchmarks.

All 5 models were imported and used from scikit-learn. We also used scikit-learn's HalvingGridSearchCV class to explore the optimal hyperparameters.

\section{Evaluation Methods}
\label{app:Eval}
\noindent \textbf{Accuracy} measures the proportion of correct predictions and is defined as $\text{Accuracy} = \frac{\text{n}_{correct}}{\text{n}_{samples}}$. Here, $\text{n}_{correct}$ is the number of correct predictions, and $\text{n}_{samples}$ is the total number of samples.
\noindent \textbf{F1 score}, a harmonic mean of Precision and Recall, is calculated as $\text{F1 score} = 2 \times \frac{\text{Precision} \times \text{Recall}}{\text{Precision} + \text{Recall}}$, where $\text{Precision} = \frac{\text{TP}}{\text{TP} + \text{FP}}$ and $\text{Recall} = \frac{\text{TP}}{\text{TP} + \text{FN}}$.

\noindent \textbf{AUC score} represents the area under the ROC curve, which plots the True Positive Rate (TPR) against the False Positive Rate (FPR) at various threshold settings.

\begin{table}[H]
\centering
\caption{Dataset Statistics}
\resizebox{0.6\columnwidth}{!}{
\begin{tabular}{l|cccl}
\hline
Dataset    & \#Variable & \#Class & \#Instance            \\ \hline
Blood~\cite{blood_transfusion_service_center}      & 4        & 2       & 583         \\
Breast Cancer~\cite{misc_breast_cancer_14}     & 31        & 2       & 569      \\
Creditcard~\cite{misc_credit_approval_27}    & 15        & 2       & 690      \\
German Credit~\cite{credit-g}       & 20        & 2       &1000+           \\
ILPD~\cite{indian_liver_patient_dataset}       & 11       & 2       & 583          \\
Loan~\cite{mazaharul_hasnine_mirza_2023}     & 10        & 2       & 615      \\
Salary~\cite{misc_census_income_20}   & 14        & 2       & 1000+      \\
Steel Plate~\cite{misc_steel_plates_faults_198}      & 34       & 2       & 1000+\\
CMC~\cite{misc_contraceptive_method_choice_30}      & 9       & 3       & 1000+\\
OGB~\cite{rabie_el_kharoua_2024_OGB}      & 13       & 3       & 1000+\\
Restaurant~\cite{rabie_el_kharoua_2024_restaurant}      & 6       & 3       & 1000+\\ 
\hline
\end{tabular}
}
\label{data_info}
\end{table}

\subsection{Selecting Pre-trained LLM}
Our research focuses not merely on prompt tuning using LLMs but on modifying the structure itself to construct a model that demonstrates high performance in classification. Specifically, one of our methodologies involves a verbalizer that requires direct access to the model's loss function and vocabulary. Therefore, we need to choose a powerful yet completely open-source LLM. Hence, we selected GPT-J 6B~\cite{gpt-j} model and LLaMA-3 8B model. Both models exhibit strong performance in inference based on extensive pre-trained knowledge and have the advantage of being fully open-source. Additionally, we further validated our approach using black-box models such as GPT-3.5.

\begin{figure}[ht!]
\centering
\includegraphics[width=1\columnwidth]{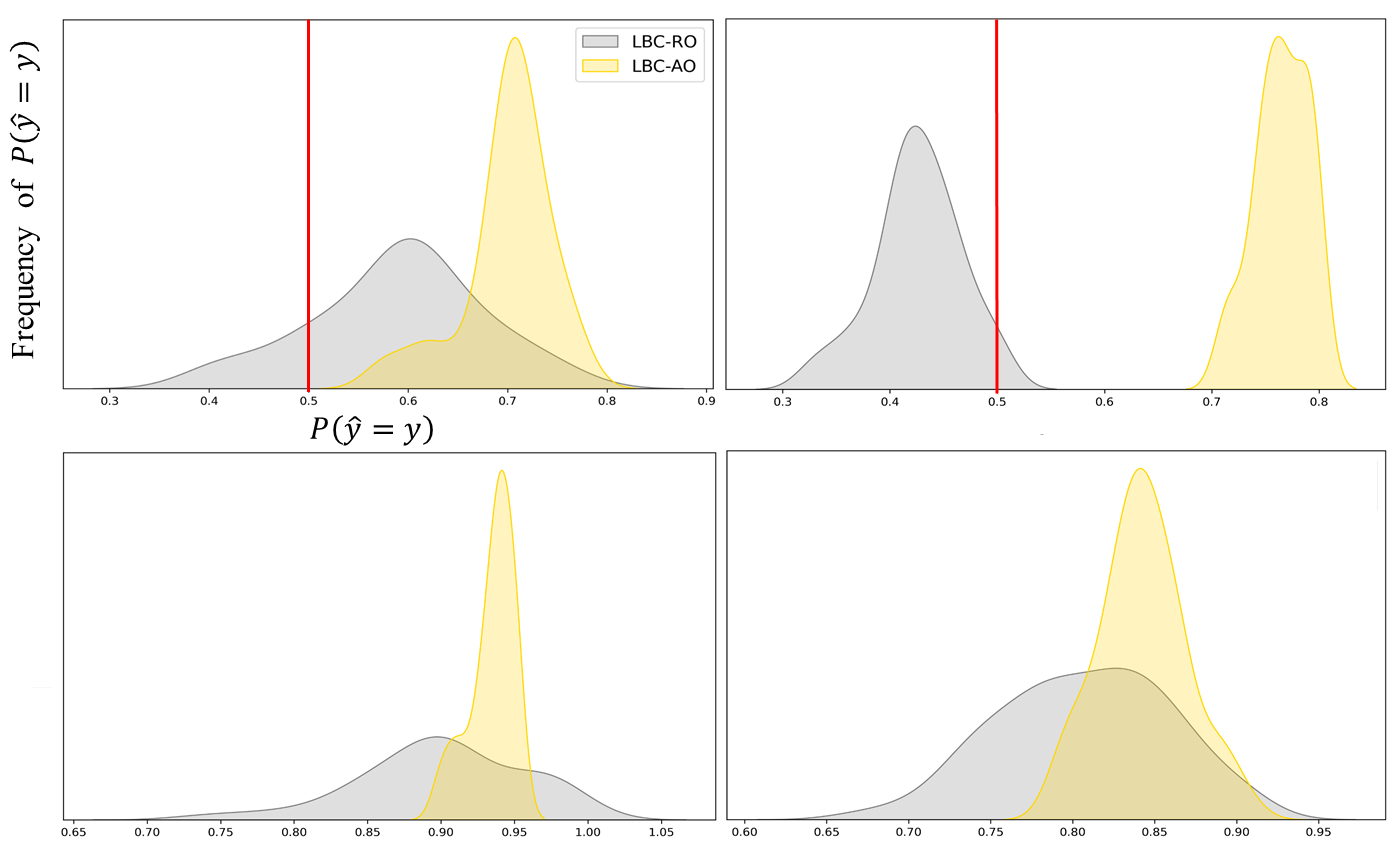}
\vspace{-0.5cm}
\caption{Frequeny of $P(\hat{y} = y)$ for the two prompt generation methods. We repeated the prompt generation 100 times for each of the four randomly selected examples from the Creditcard dataset in two ways: random order (RO) and advanced order (AO). The horizontal axis represents the model's probability for the correct class $y$, and the vertical axis shows frequency. The AO method provides more consistent and accurate results than RO. The red vertical line indicates the prediction boundary, where the differences between the two methods lead to varied predictions.}
\end{figure}

\end{document}